\documentclass[sigconf]{acmart}

\usepackage{booktabs} 

\setcopyright{rightsretained}

\editor{Chetan Arora}
\editor{Parag Chaudhuri}
\editor{Subhransu Maji}

\acmArticle{03}

\copyrightyear{2021}
\acmYear{2021}
\setcopyright{acmcopyright}\acmConference[ICVGIP '21]{Indian Conference on Computer Vision, Graphics and Image Processing}{December 19--22, 2021}{Jodhpur, India}
\acmBooktitle{Indian Conference on Computer Vision, Graphics and Image Processing (ICVGIP '21), December 19--22, 2021, Jodhpur, India}
\acmPrice{15.00}
\acmDOI{10.1145/3490035.3490260}
\acmISBN{978-1-4503-7596-2/21/12}

\begin{document}

\title{MERANet: Facial Micro-Expression Recognition using 3D Residual Attention Network}

\author{Viswanatha Reddy Gajjala}
\affiliation{%
  \institution{IIIT Sri City}
  \city{Chittoor}
  \country{India}}
\email{viswanatha.g15@iiits.in}

\author{Sai Prasanna Teja Reddy}
\affiliation{%
  \institution{The University of Chicago}
  \city{Chicago}
  \country{USA}}
\email{bogireddyteja@uchicago.edu}

\author{Snehasis Mukherjee}
\affiliation{%
  \institution{Shiv Nadar University}
  \city{Greater Noida}
  \country{India}}
\email{snehasis.mukherjee@snu.edu.in}

\author{Shiv Ram Dubey}
\affiliation{%
  \institution{IIIT Sri City}
  \city{Chittoor}
  \country{India}}
\email{srdubey@iiits.in}

\begin{abstract}
Micro-expression has emerged as a promising modality in affective computing due to its high objectivity in emotion detection. Despite the higher recognition accuracy provided by the deep learning models, there are still significant scope for improvements in micro-expression recognition techniques. The presence of micro-expressions in small-local regions of the face, as well as the limited size of available databases, continue to limit the accuracy in recognizing micro-expressions. In this work, we propose a facial micro-expression recognition model using 3D residual attention network named MERANet to tackle such challenges. The proposed model takes advantage of spatial-temporal attention and channel attention together, to learn deeper fine-grained subtle features for classification of emotions. Further, the proposed model encompasses both spatial and temporal information simultaneously using the 3D kernels and residual connections. Moreover, the channel features and spatio-temporal features are re-calibrated using the channel and spatio-temporal attentions, respectively in each residual module. Our attention mechanism enables the model to learn to focus on different facial areas of interest. The experiments are conducted on benchmark facial micro-expression datasets. A superior performance is observed as compared to the state-of-the-art for facial micro-expression recognition on benchmark data.

\end{abstract}


\maketitle

\begin{figure}[!t]
\includegraphics[width=0.48\textwidth]{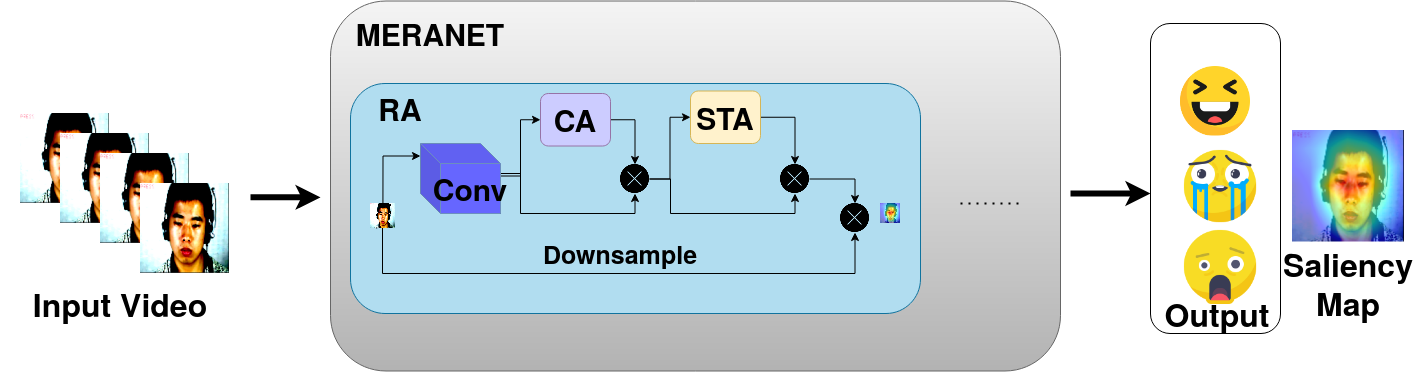}
\caption{An overview of the proposed MERANet model. In the figure: RA corresponds to Residual Attention (RA) Module which consists of channel attention (CA) and spatio-temporal attention (STA). The MERANet model is comprised of multiple stacked RA modules. Saliency map provided by the MERA Net model is shown at the right side of the figure.}
\label{block_diagram}
\end{figure}

\section{Introduction}
In recent years, facial expression recognition using deep learning techniques have gained lot of popularity among the research community because of its potential applications across various fields, such as psychology, marketing, security, etc. However, recognizing true human emotions from facial expressions often becomes unreliable due to the performer's deliberate control on the expressions, depending on the social circumstances. Hence, micro-expression recognition has evolved as an attempt to capture the true emotion of a performer \cite{ekman}. Micro-expressions are involuntarily exposed short term expressions that appear during an active suppression of facial expression to hide the true emotion. Micro-expressions are very short spanned (fraction of a second; usually around 0.5 seconds), which makes the task of capturing the relevant features for recognizing emotion as challenging \cite{ekman1997face}.

Efforts have been made to apply Convolutional Neural Networks (CNNs) for micro-expression recognition \cite{patel2016selective,takalkar2017}. Although CNN models are able to automatically capture useful features from the videos, but they are often unable to capture the minute textural changes on various parts of the face, observed during micro-expressions. To capture the minute textural information, facial landmarks are used for feature extraction \cite{grobova2017}. However, even the temporal features play an important role in recognising facial micro-expressions.

A few efforts have also been made in capturing temporal cues from optical flow, for recognizing micro-expressions \cite{li2018,liong2018,li2018flow}. However, features based on optical flow are frequently hampered by unnecessary motion information from background pixels. Another line of thought for capturing temporal features suggests applying a Long Short Term Memory (LSTM) on the spatial features \cite{kim2017, kim2016micro, khor2018}. A few 3D CNN based models can also be found in the literature, to capture temporal cues \cite{reddy2019,liong2019}. The existing approaches of applying temporal features often fails to capture the minute texture information from micro-expression videos, where the expression continues for the very short span of time. Attention based models often show efficacy in extracting the minute texture information \cite{humanperception,hinton}. However, attention mechanism is still not properly explored for micro-expression recognition task. Wang et al. proposed a spatial attention model for micro-expression recognition \cite{wang2019}. Wang et al. proposed an attention mechanism for recognizing micro-expression using just the apex frame from the micro-expression video. The model completely ignores the temporal information between the frames. Whereas, the proposed model considers both spatial and temporal information simultaneously by applying 3D Conv operations while recalibrating features using attention mechanism and residual connections to learn more fine grained information. To capture the temporal information from the short span of video, where the micro-expression takes place, we propose a 3D channel and spatio-temporal attentions based model called as MERANet. Due to the utilization of the 3D Spatial attention and Channel attention, the proposed model is able to ignore the background information while focusing on the region of interest precisely which is consistent across the frames of the video and responsible for the given micro expression. Fig. \ref{block_diagram} represents the overall idea of the proposed MERANet model. The main contributions of this paper are as follows:
\begin{itemize}
\item A 3D residual attention is introduced in this paper for recognizing subtle features related to micro-expressions by attention guided re-calibration of features.

\item In order to prioritize the important channels and spatio-temporal features, the channel attention and spatio-temporal attention blocks are used as part of the residual module.

\item The increase in number of parameters and FLOPs in the proposed model is negligible as compared to the vanilla model.

\item Extensive experiments are performed to show the impact of the proposed 3D residual attention model on micro-expression recognition. We show the saliency map to visualize the effect of the proposed approach in extracting features from faces during expression.

\end{itemize}

Next we provide a survey of literature for facial micro-expression recognition. 

\begin{figure*}[!t]
\includegraphics[width=\textwidth]{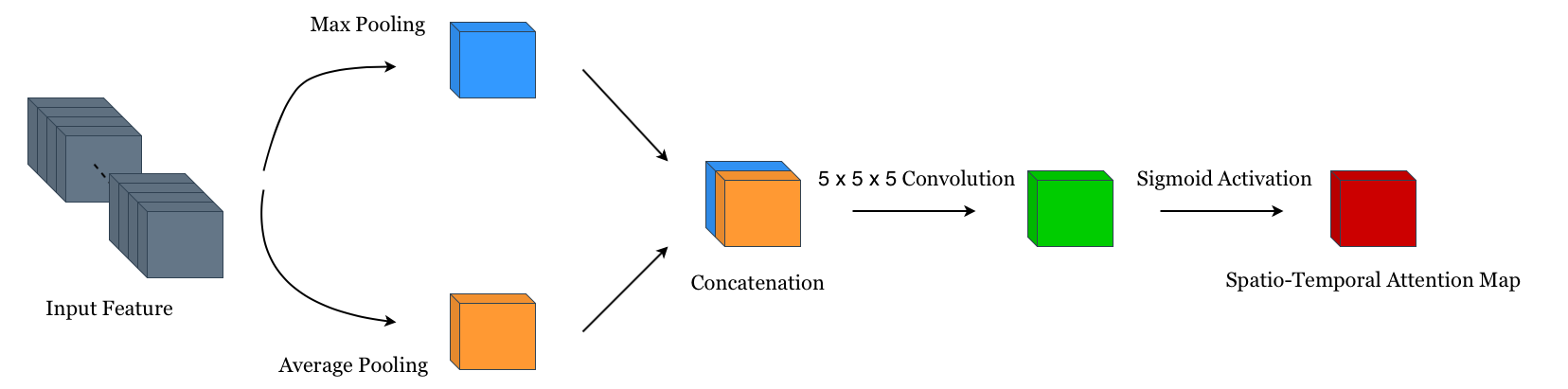}
\caption{The proposed Spatial-Temporal Attention block.}
\label{ST_attention}
\end{figure*}

\section{Literature Review}
The proposed method aims to recognize the micro-expressions using attention models. This section provides a discussion on the literature related to micro-expression recognition, followed by a discussion on how attention models are applied to solve various problems in Computer Vision.
\subsection{Facial Micro-Expression Recognition}
The facial micro-expression recognition methods can be categorized into two categories: handcrafted feature based methods and deep learning based methods.

\subsubsection{Handcrafted Feature based Methods}
The traditional approaches for micro-expression recognition were mostly focused on superficial attributes involving complex calculations. Polikovsky \cite{polikovsky2009facial} proposed a three step algorithm that divides the facial region into twelve parts to obtain facial cubes. They computed 3D orientation histogram descriptor to capture correlations across the frames for each facial cube. They classified the expression based on the normalized descriptor vectors. Shreve et al. \cite{shreve2011macro} utilized the facial strain caused by the non-rigid motion of facial muscles during an expression. The magnitude of strain was found using optical flow over different regions of the face. Pfister et al. \cite{pfister2011recognising} used temporal interpolation model to address variable video length problem. They introduced spatiotemporal local texture descriptors to generate feature vector and finally, various kinds of classifiers have been tried to classify the micro-expression. Huang et al. \cite{huang2016spontaneous} proposed a spatiotemporal completed local quantized pattern that extracts sign, magnitude and orientation components from an expression and constructed a codebook for each component that holds more dynamic pattern representations. Liu et al. \cite{liu2016main} proposed a method that generates a compact feature vector by finding main directional mean optical flow using optical flow field over different regions of interest of the face. The final feature vector is fed into an SVM to classify micro-expressions. Zhao et al. \cite{zhao2007dynamic} proposed a computationally compact feature using Local Binary Pattern - Three Orthogonal Planes (LBP-TOP) which can efficiently extract co-occurring features from neighboring points. Lu et al. \cite{lu2018} proposed a fusion based feature that finds differentials of optical flow over horizontal and vertical components. The fused feature vector is fed into an SVM classifier for classification. 

\subsubsection{Deep Learning based Methods}
In most of the computer vision applications, the deep learning techniques outperform the handcrafted feature based methods due to its increased feature learning capability from the data \cite{lecun2015deep}. Patel et al. \cite{patel2016selective} proposed a deep learning model for micro-expression recognition, where they used transfer learning to initialize the CNN with weights of a CNN trained on largely available macro-expression data to avoid overfitting. Takalkar implemented a variety of data augmentation techniques to generate synthetic data, resembling different environments to deal with lack of training data. Finally a CNN is applied to classify micro-expressions \cite{takalkar2017}. Qiuyu et al. \cite{li2018} proposed a three step algorithm where a multi-task CNN was used to find the facial landmarks. Deep CNN was used to calibrate the optical flow features from different facial regions. Finally, an SVM was employed to classify the micro-expressions. Liong et al. \cite{liong2018} collected the optical flow feature for each micro-expression and used it as input to a CNN for classification. Li et al. \cite{li2018flow} proposed a 3D flow based CNN that uses gray scale images along with components of optical flow in different directions as input to classify micro-expressions. Wang et al. \cite{wang2019} proposed an image based residual model that uses micro-attention units and transfer learning to classify micro-expressions. The model fails to consider temporal information across the frames, which is important for micro-expression videos. Several proposals can be found in the literature \cite{kim2017, kim2016micro, khor2018} that used two step architecture where, a CNN is applied on each frame of the micro-expression video to generate spatial features and then a Long Short Term Memory (LSTM) based Recurrent Neural Network (RNN) is used to learn the temporal inter-dependencies between different spatial features from previous step. Reddy et al. \cite{reddy2019} proposed a 3D CNN based model that tries to capture both spatial and temporal information in a micro-expression video simultaneous using 3D kernels. Liong et al. \cite{liong2019} proposed a shallow triple stream 3D CNN that uses optical strain, horizontal and vertical flows from the onset and apex frames to classify micro-expressions.

The existing methods focus neither on the minute spatial changes on face regions during micro-expressions nor on the temporal cues, as the micro-expressions last for a small span of time. A spatial attention mechanism can help in capturing the minute spatial changes in facial regions during micro-expression. In addition to that, a temporal attention can help in focusing on the smaller span of the micro-expression.

\subsection{Attention Mechanism in Literature}
The process of human perceptual analysis is based on the attention mechanism \cite{humanperception}. Human vision obtains the target area within the scene, that needs to be focused on, and pays more attention on the target area to obtain more detailed information about the scene that helps to capture the visual structure of the target better \cite{hinton}. Attention mechanism is used to find the region of interest within an image and to highlight the representation of the region of interest \cite{cvatt}. The two main aspects of attention mechanism in deep learning are (a) to obtain the meaningful channels for the respective input feature and (b) to pay attention to the most informative channels to obtain the salient locations. 

Recently, several studies attempted to apply attention mechanism to enhance the performance of CNNs in a range of visual tasks (especially the tasks where objects play vital roles), such as image classification, image localization and video understanding \cite{topdown,Transformer}. Wang et al. \cite{RAN} proposed a Residual Attention Network (RAN) which uses a trunk-and-mask module to incorporate an attention mechanism. By re-weighting the feature map, the network not only performs well but is also robust to input noise. Hu et al. \cite{squeezeexcite} introduced a Squeeze-and-Excitation module to exploit the inter-channel relationship. Adding to this, Woo et al. \cite{CBAM} introduced spatial attention to further improve the image classification performance. In \cite{gesture}, Peng et al. designed an attention branch that helps to spotlight finger micro-gesture and reduces the noise introduced from the background and wrist. The spatial attention has improved the recognition accuracy by $3\%$. However the attention block proposed in the study of \cite{gesture}, increased the parameters by about $2M$. Similarly, for micro-expression recognition, an attention mechanism helps the model to focus on the face and important facial regions, and suppress the irrelevant facial areas and background.

The existing attention based mechanisms have not yet utilized the 3D channel and spatio-temporal attentions, which are important in the context of facial micro-expression recognition. In this work, we design a novel and effective 3D residual attention mechanism to simultaneously learn fine and subtle features along spatial, temporal and channel dimensions. The proposed attention modules, including channel attention and spatio-temporal attention, add just a few parameters to the base model, so that the increase in computation overhead is negligible. Next we illustrate the proposed method for micro-expression recognition.

\section{Proposed MERANet Model for Facial Micro-Expression Recognition}
Given a facial micro-expression video, our objective is to obtain a good representation in terms of their emotional score. We propose a deep learning model, named MERANet, for micro-expression recognition using 3D residual attention network. The proposed residual attention (RA) network is constructed by stacking up multiple RA blocks. The RA block comprises of two major sub-modules: (a) A Spatio-Temporal Attention module to extract the spatio-temporal importance map and (b) A Channel Attention module to extract more discriminative features by prioritizing the important channels.

\subsection{Spatio-Temporal Attention Block}
We compute the spatio-temporal attention map by utilizing the spatial-temporal relationship among the features. The spatio-temporal attention focuses on where we need to pay more attention in the input feature volume.
\begin{figure*}[!t]
\includegraphics[width=\textwidth]{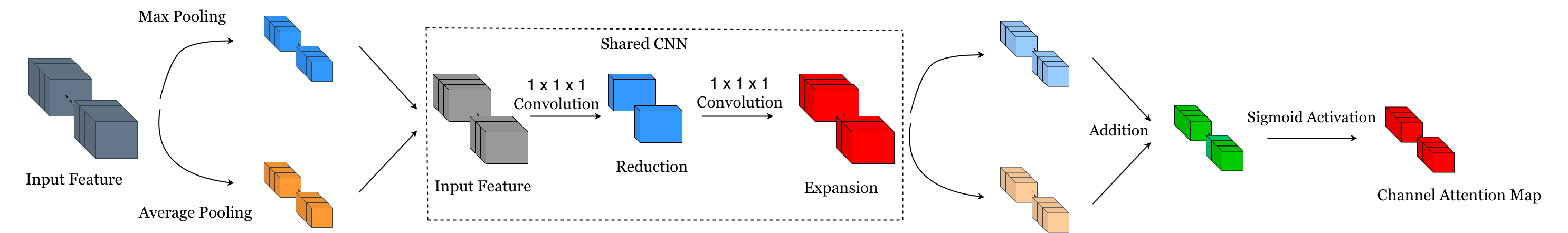}
\caption{The proposed Channel Attention block. We shrink the spatial dimensions of the input feature map to 1 x 1 x 1 (height, width, depth) with C channels. The number of channels are unaltered. The number of channels is always greater than one ( C >> 1).}
\label{channel_attention}
\end{figure*}

In order to compute the spatio-temporal attention without increasing the model parameters, we first apply average-pooling and max-pooling operations along the channel axis and aggregate them by concatenating to generate an efficient feature descriptor $\mathcal{F}^{st} \in \mathcal{R}^{2 \times T' \times W' \times H'}$, where $T'$, $W'$ and $H'$ represent the temporal dimension, width, and height respectively, of the generated feature descriptor. The pooling operation is employed along the channel axis as it is effective in highlighting the informative regions \cite{payattention}. 

\noindent The average pooling across the channels is defined as,
\begin{equation}
\mathcal{F}_{avg}^{st}(1, t, h, w) = 
\frac{1}{C'}\sum_{c=1}^{C'}{F}_{c, t, h, w},
\label{eq_st_average}
\end{equation}

\noindent and the max pooling across the channels is defined as,
\begin{equation}
\mathcal{F}_{max}^{st}(1, t, h, w) 
= \max_{c \in \{1,2,...,C'\}} (F_{c, t, h, w}),
\label{eq_st_max}
\end{equation}
where $t \in \{1,2,..., T'\}$, $h \in \{1,2,..., H'\}$, and $w \in \{1,2,..., W'\}$, $F \in \mathcal{R}^{C' \times T' \times W' \times H'}$ corresponds to the input feature volume and $C'$ is the  number of channels in the input feature ($F$).

In order to limit the module complexity, a single convolution operation is performed on the feature descriptor to produce a more refined feature map followed by a sigmoid activation function. The final generated attention importance map ($\mathcal{A}^{st}$) learns the importance score for each spatial-temporal location. The $\mathcal{A}^{st}$ is defined as,
\begin{equation}
    \mathcal{A}^{st}  = \sigma[ W_{5 \times 5 \times 5} (\mathcal{F}^{st})],
    \label{eq_st_attention}
\end{equation}
where $\mathcal{F}^{st}$ is given as, 
\begin{equation}
    \mathcal{F}^{st} = Concat(\mathcal{F}_{avg}^{st} ; \mathcal{F}_{max}^{st}),
\end{equation}
and $Concat(\mathcal{F}_{avg}^{st} ; \mathcal{F}_{max}^{st}) \in \mathcal{R}^{2 \times T' \times W' \times H'}$ is the concatenated features of average pooling and max pooling across the channel dimension, $\mathcal{A}^{st} \in \mathcal{R}^{1 \times T' \times W' \times H'}$ is the generated attention importance map, $\sigma$ denotes the sigmoid function and $W_{5 \times 5 \times 5}$ corresponds to a convolution operation with the kernel of size $5$ with \textit{same} padding. Note that all the values in the importance map are in the range 0 to 1, thus mimicking the probability of being important. Basically, $\mathcal{A}^{st}$ encodes that which spatio-temporal information to emphasize or suppress. The block diagram of the spatio-temporal attention module is shown in Fig. \ref{ST_attention}. Element-wise multiplication is employed between the input feature volume and the spatio-temporal importance map ($\mathcal{A}^{st}$) to re-weight each pixel value and obtain the refined feature map. The re-calibrated feature activation ($F^{st}$) is defined as,
\begin{equation}
    F_{c}^{st}|_{\forall c \in \{1,2,...,C'\}} = F_c \otimes \mathcal{A}^{st},
    \label{eq_st_recalibration}
\end{equation}
where $F_c$ and $F^{st}_c$ are the $c^{th}$ channel of input and re-calibrated output using spatio-temporal attention and $\otimes$ denotes the element-wise multiplication. 

\begin{figure*}[!t]
\includegraphics[width=\textwidth]{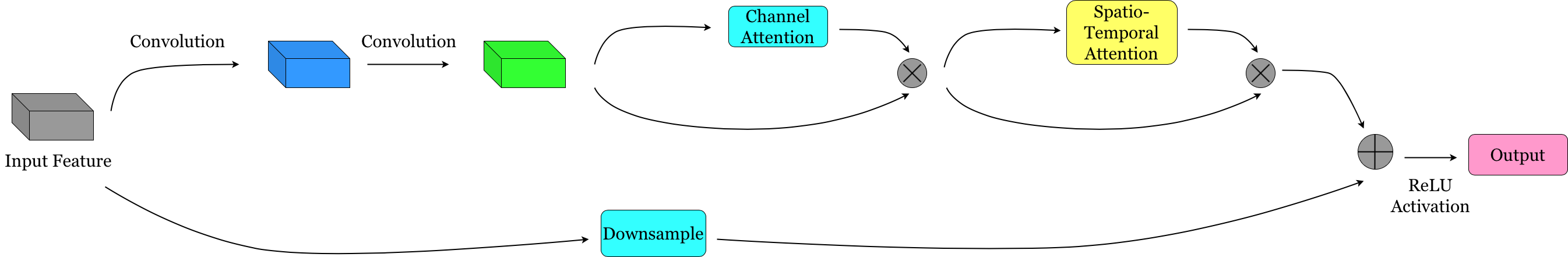}
\caption{The proposed 3D RA module by integrating the 3D convolutions, channel attention and spatio-temporal attention blocks. Here `+' stands for element wise addition, `X' stands for element wise multiplication and the output corresponds to a recalibrated residual feature map that goes through channel and spatial attention sequentially.}
\label{ra_block}
\end{figure*}

\subsection{Channel Attention Block}
The spatio-temporal attention captures the weight for each spatio-temporal feature volume. However, it is not able to distinguish the weight among the different channels of feature volume. Thus, we also propose a channel attention block in the residual framework. The channel features in the CNNs are exploited extensively to extract more discriminative features. We compute a channel attention map by utilizing the inter-channel relationship of features. The channel attention map focuses on finding the meaningful channels for the respective input feature volume. Integrating a channel attention map helps to improve the learning ability of the model by re-calibrating the channels of the input feature volume with increased or decreased weight. In order to capture the channel attention effectively for a feature volume ($\mathcal{F^{'}} \in \mathcal{R} ^{ C^{'} \times T^{'} \times W^{'} \times H^{'}}$), first we squeeze the spatial and temporal dimensions of the input feature volume by applying average pooling and max pooling and then we combine the squeezed features by concatenating them. 

\noindent The average pooling over spatial and temporal dimensions is given as,
\begin{equation}
\mathcal{F}_{avg}^{ch}(c,1,1,1)= \frac{1}{T' \times W' \times H'}  {\sum_{t=1}^{T'} \sum_{h=1}^{H'} \sum_{w=1}^{W'} \mathcal{F'}(c, t, h, w)},
\label{eq_channel_average}
\end{equation}
and the max pooling over spatial and temporal dimensions are given as,
\begin{equation}
\mathcal{F}_{max}^{ch}(c,1,1,1) = \max_t \max_h \max_w \mathcal{F'}(c, t, h, w),
\label{eq_channel_max}
\end{equation}
where $t \in \{1,2,...,T'\}$, $h \in \{1,2,...,H'\}$, $w \in \{1,2,...,W'\}$, $\mathcal{F}_{avg}^{c}  \in \mathcal{R}^{C' \times 1 \times 1 \times 1}$, $\mathcal{F}_{max}^{c} \in \mathcal{R}^{C' \times 1 \times 1 \times 1}$, $c \in \{1, 2, ..., C'\}$, and $C^{'}$ is the number of channels.

The channel descriptors $\mathcal{F}_{max}^{ch}$ and $\mathcal{F}_{avg}^{ch}$ are then fed to a shared sub-network to fully capture the inter-dependencies. We propose two design choices for the shared sub-network. One variant is with the convolutional layers and other with the multi-layer perceptron layers (MLP). Our experimental results show that the shared network with convolutions is better than the MLP. The detailed analysis will follow in experimental results in Section 4. The shared convolutional network consists of a convolution layer with the channel reduction mechanism (through $1 \times 1 \times 1$ convolution) to reduce the computation overhead. The ReLU activation is used to introduce the non-linearity followed by a dimension increasing layer to match the input channel dimension (through $1 \times 1 \times 1$ convolution).

The final channel attention map ($\mathcal{A}^{ch}$) is computed as,
\begin{equation}
    \mathcal{A}^{ch} = \sigma[ W_{e} (\varphi (W_{s}(\mathcal{F}_{avg}^{ch})) )  + W_{e} (\varphi (W_{s}(\mathcal{F}_{max}^{ch})) )],
    \label{eq_ch_attention}
\end{equation}
where $\sigma$ and $\varphi$ correspond to the Sigmoid and ReLU activation functions, respectively, $W_{s}$ and $W_{e}$ are the convolution operations in the shared network with $W_s(.) \in \mathcal{R}^{{C' \over r} \times 1 \times 1 \times 1} $ and $W_e(.) \in \mathcal{R}^{C' \times 1 \times 1 \times 1}$, $r$ is the channel reduction ratio in the shared sub-network, and $\mathcal{A}^{ch}  \in \mathcal{R}^{C' \times 1 \times 1 \times 1}$. The reduction ratio($r$) is set to 16 in this paper.

Note that all the values in the channel importance map are in the range $0$ to $1$. Thus, it represents the probability of being important for each channel. 
The block diagram of the channel attention module is shown in Fig. \ref{channel_attention}.

Channel-wise multiplication is employed between the input feature volume and the channel importance map ($\mathcal{A}^{c}$) to obtain the re-calibrated channels of the output feature map. The re-calibrated feature activation ($F'_{ch}$) is computed as,
\begin{equation}
    F'_{ch}(c,t,h,w) = F'(c,t,h,w) \otimes \mathcal{A}^{ch}(c,1,1,1),
    \label{eq_ch_recalibration}
\end{equation}
for $\forall c \in \{1,2,...,C'\}$, $\forall t \in \{1,2,...,T'\}$, $\forall h \in \{1,2,...,H'\}$, and $\forall w \in \{1,2,...,W'\}$,
where $F'$ and $F'_{ch}$ are the input and re-calibrated output using channel attention, respectively. 

\subsection{Residual Attention Module}
In this paper, we propose a 3D RA module by utilizing the spatio-temporal and channel attention blocks. In contrast to usual residual block, we introduce the attention blocks into the residual module. The residual attention module comprises of two 3D convolution operations, channel attention, spatial-temporal attention, residual connection followed by an activation function. For simplicity, we ignore the batch-norm and activation layers in the block diagram. 

The block diagram for the proposed 3D RA module is shown in Fig. \ref{ra_block}. Suppose $\mathcal{I} \in \mathcal{R}^{C \times T \times W \times H}$ is the input feature volume to the proposed residual attention module and $\mathcal{I}^{conv} \in \mathcal{R}^{C' \times T' \times W' \times H'}$ is the generated intermediate feature volume after two convolution blocks. We apply the channel attention operation over $\mathcal{I}^{conv}$ to generate the channel attention map $\mathcal{A}^{ch}$ as defined in Eq. (\ref{eq_ch_attention}) which uses the average pooling and max pooling over spatial and temporal dimensions as given in Eq. (\ref{eq_channel_average}) and (\ref{eq_channel_max}), respectively. The re-calibrated intermediate feature after channel attention is given using Eq. (\ref{eq_ch_recalibration}) as,
\begin{equation}
   \mathcal{I}^{ch} = \mathcal{I}^{conv} \otimes \mathcal{A}^{ch}.
\end{equation}
Next, we apply the spatio-temporal attention operation over $\mathcal{I}^{ch}$ to generate the spatio-temporal attention map $\mathcal{A}^{st}$ as defined in Eq. (\ref{eq_st_attention}) which uses the average pooling and max pooling across channel dimension as given in Eq. (\ref{eq_st_average}) and (\ref{eq_st_max}), respectively. The re-calibrated intermediate feature after spatio-temporal attention is given using Eq. (\ref{eq_st_recalibration}) as,
\begin{equation}
   \mathcal{I}^{st} = \mathcal{I}^{ch} \otimes \mathcal{A}^{st}.
\end{equation}
The channel attention excites the important channels separately, whereas the spatio-temporal attention excites the important spatio-temporal features. The last step in the proposed residual module is to add the generated re-calibrated features with the original input of this block. The final output of the proposed residual module ($\mathcal{O} \in \mathcal{R}^{C' \times T' \times W' \times H'}$) is computed as,
\begin{equation}
    \mathcal{O} = \varphi[\mathcal{I}^{st} + d(\mathcal{I})]
\end{equation}
where $\varphi$ is the ReLU activation function and $d$ is a down-sampling function to change the dimension of input from ${C \times T \times W \times H}$ to ${C' \times T' \times W' \times H'}$.
The down-sampling function is used in the residual path to match the dimensions of the features $\mathcal{I}^{st}$ and $\mathcal{I}$.  
    (a) If the dimensions of the $\mathcal{I}^{st}$  and $\mathcal{I}$ are same then the down-sampling will be a simple identity mapping with no additional parameters added.
    (b) If the dimensions of the $\mathcal{I}^{st}$  and $\mathcal{I}$ are different then the down-sampling function will use $1 \times 1 \times 1$ convolution with the stride $2$ to match the dimensions.





\subsection{Network Architecture}
The residual network (ResNet) and its variants are one of the most successful architectures for image classification~\cite{ResNet}~\cite{ResNeXt}~\cite{wideresnet}. A basic residual module consists of two convolutional layers in a sequence. The first convolutional layer is followed by a batch normalization and a ReLU activation, whereas the second convolution layer is followed by a batch normalization. The residual module adds the output of the second convolution layer to the input which is followed by a ReLU activation function. 
The ResNet uses the residual connections that enforces the network to learn the difference of transformation instead of complete transformation.
It facilitates better gradient flow during the back-propagation and helps in the optimization of the network. Thus, facilitating the training of very deep networks feasible. Due to the residual property, the ResNet abates the vanishing gradient problem. 

We also generate the proposed MERANet architecture by following the ResNet architecture in this work.
The MERANet-18 is constructed by stacking up the multiple 3D residual attention modules. We use the identity connections and zero padding in the residual paths of the residual attention module to avoid increasing the number of parameters of the model. The 3D residual attention modules are used as the basic building blocks of the proposed MERANet model. 
We add a stem network to get the reasonable size feature map similar to the ResNet. We also employ a fully connected layer at the end to classify the emotion score. The detailed architecture is summarized in supplementary.
 
\section{Experimental Settings}
In this section, first we provide a brief description of four benchmark micro-expression datasets used for the experiments, followed by the pre-processing details and the training settings.

\subsection{Micro-Expression Datasets Used}
We have used four benchmark micro-expression datasets to evaluate the proposed model, namely CASMEI, CASMEII, CAS(ME)$^{2}$ and SAMM. 

CASME I Dataset \cite{casme}:
CASME I dataset is divided into two parts: CASME A and CASME B, which are constructed under two different conditions. CASME A contains images with $1280 \times 720$ resolution under natural light settings. Whereas, CASME B contains images with resolution of $640 \times 480$ under the artificial light settings. Both settings have a temporal resolution of $60$ fps. Altogether, CASME dataset contains $195$ micro-expressions belonging to seven different classes. For this experiment, we have chosen three classes, including disgust ($34$ samples), repression ($36$ samples), and tense ($45$ samples) based on the higher number of samples in these classes.

CASME II Dataset \cite{Yan2014CASMEIA}:
CASME II is a spontaneous micro-expressions dataset built under well-controlled laboratory environment. The samples have a higher face resolution at $280 \times 340$ pixels and a relatively high temporal resolution ($200$ fps). The database has five micro-expression categories. The number of samples for each category is distributed unequally (there are $60$ samples for disgust and only $7$ samples for sadness). Since the data is not uniformly distributed across the classes to obtain consistent results, we have used three classes, including disgust ($63$ samples), happy ($32$ samples), and surprise ($28$ samples) based on the higher number of samples in these classes. 
CAS(ME)$^2$ Dataset \cite{casmesquare}: CAS(ME)$^2$ dataset comprises of both macro-expressions and micro-expressions. This dataset is constructed by showing $9$ elicitation videos to $22$ participants under artificial light settings. The elicitation videos contain three emotions (happy, angry, and disgust) evoking videos. CAS(ME)$^2$ contains the samples with $640 \times 480$ resolution at $30$ frames per second. In our experiment, we have only used micro-expressions of CAS(ME)$^2$ dataset. The used dataset contains three expressions, namely happy ($18$ samples), angry ($22$ samples), and disgust ($17$ samples). 

SAMM Dataset \cite{davison2018}: SAMM dataset comprises of 159 samples with $2040 \times 1088$ resolution at $200$ frames per second. The database has seven micro-expression categories. The number of samples for each category is distributed unequally (there are $57$ samples for anger and just $6$ samples for sadness). Since the data is not evenly distributed across the classes to obtain consistent results, we have used the following three classes: anger ($57$ samples), happiness ($26$ samples), and other ($26$ samples). We have not included rest of the 4 classes as the number of samples in each of these classes is too less ($\leq 15$) to train a deep learning system. 

\subsection{Pre-Processing}

In the experiments, first, we select the apex frame in the video corresponding to the micro-expression in order to generate a training sample. As the data is captured at different frame rates per second, we have assigned temporal depth for SAMM as 64, CASMEI and CASMEII as 32 and CAS(ME)$^{2}$ as 16. Also, we have considered samples having atleast 16 frames for CASMEI and CASMEII in this experimentation. A $16$-frame clip is generated around the selected apex frame. If the video is shorter than $16$ frames, then we replicate the edge frames as many times as necessary. The temporal dimension is selected based on the average number of frames available in a video clip. We spatially resize the sample to $112 \times 112 pixels$. The size of each sample is $3$ channels $\times$ $16$ frames $\times$ $112$ pixels $\times$ $112$ pixels, and each sample is horizontally flipped with $0.5$ probability. Mean subtraction is applied for each sample by subtracting the mean values from the sample for each channel. Each sample is normalized by the standard deviation.


\begin{table*}[!t]
\centering
\caption{The performance of the proposed MERANet model in terms of the facial micro-expression recognition accuracy as compared to the existing state-of-the-art models. Note that the results of existing methods are taken from the respective papers. The HCM and DLM denote the hand crafted method and deep learning method, respectively.}
\begin{tabular}{c|c|c|c|c|c c c}
\hline
\textbf{Method} &
\textbf{CASME I} & \textbf{CASME II} & \textbf{CAS(ME)$\widehat{}^2$} & \textbf{Year} & \textbf{Method Type} \\
\hline
STCLQP \cite{huang2016spontaneous} & 57.31 $\%$ & 58.39 $\%$ & - & 2016 & HCM \\ 
\hline
FMBH \cite{lu2018} & 61.33 $\%$ & 69.11 $\%$ & 73.67 $\%$ & 2018 & HCM \\
\hline
CNN + Optical Flow \cite{li2018} & 56.60 $\%$ & 56.94 $\%$ & - & 2018 & DLM \\ 
\hline
3D Flow CNN \cite{li2018flow} & 54.44 $\%$ & 59.11 $\%$ & - & 2018 & DLM \\
\hline
LEARNet \cite{Verma_2020} & 80.62 $\%$ & 76.57 $\%$ & 76.57 $\%$ & 2019 & DLM\\
\hline
MicroExpSTCNN \cite{reddy2019} & - & - & 87.80 $\%$ & 2019 & DLM\\
\hline
MicroAttention \cite{wang2019} & - & 65.90 $\%$ & - & 2020 & DLM\\
\hline
3D Residual Network (Ours) & 81.0$\%$ & 83.3 $\%$ & 83.3 $\%$ & --- & DLM\\
\hline
MERANet (Proposed Model) & \textbf{90.5 \%} & \textbf{93.5 \%} & \textbf{91.7 \%} & --- & DLM\\
\hline
\end{tabular}
\label{table:comparison}
\end{table*}

\subsection{Training Settings}
We have performed $75\%$-$25\%$ train-test split for CASME I and CASME II and $80\%$-$20\%$ train-test split for CAS(ME)$\widehat{}^2$ due to less samples. Among the splits $80\%$ of data is used as a training set and the rest is used as a validation set. Stratified sampling is used for the validation split to maintain consistency across the results. The train-validation split-up is performed once and then the same training and validation sets are used for all the experiments.

The weights of both convolutional and fully-connected layers are initialized with the Xavier initialization proposed in \cite{Xavier}. 
We set the parameter to random values uniformly drawn from [$-rv$, $rv$], where $rv$ is defined as, $rv = \sqrt{6 \over (d_{in} + d_{out})}$ where $d_{in}$ and $d_{out}$ are the size of input and output channels, respectively. All biases are initialized to $0$. For batch normalization layers, weights are initialized to $1$.

The optimizer proposed in \cite{Uwarmup} is used for training. We use Categorical Cross Entropy as the loss function. Each model is trained for $100$ epochs with a batch size of $8$. The learning rate is initialized to $0.1$. Learning rate adjustment is one of the crucial steps while training. We use the cosine annealing strategy proposed in \cite{Cosine}. The cosine annealing decreases the learning rate from the initial value to $0$ by following the cosine function. The intuition behind using annealing is that it helps to traverse quickly from the initial parameters to a range of good parameter values with a smaller learning rate. Thus, we can explore the deeper and narrower parts of the loss function, which potentially improves the training progress by avoiding the divergence.

We avoid the overfitting problem in two ways, 1) by using the Horizontal flip data augmentation and 2) by incorporating the warm-up scheduling along with the cosine annealing for ADAM optimizer \cite{Uwarmup}, which helps significantly in the convergence. Next we discuss the results obtained by experimenting on the proposed method.

\begin{figure}[!t]
\centering
\includegraphics[width=\columnwidth]{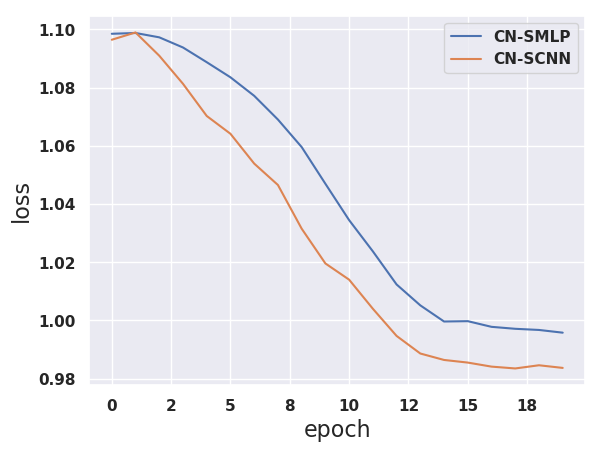}
\caption{Validation loss curves for the shared network choices, including channel attention using shared multi-layer perceptron (CN-SMLP) and channel attention using shared convolutional neural network (CN-SCNN).}
\label{fig:validationloss}
\end{figure}

\begin{table}
\centering
\caption{The performance of the proposed MERANet model in terms of the accuracy, precision, recall and F1 score.}
\begin{tabular}{c|c|c|c|c}
\hline
\textbf{Dataset} & \textbf{Accuracy} &
\textbf{Precision} & \textbf{Recall} & \textbf{F1}  \\
\hline
CASME I & 90.5$\%$ & 0.917 & 0.907  & 0.907 \\
\hline
CASME II & 93.5$\%$ & 0.94 & 0.94 & 0.933 \\
\hline
CAS(ME)$\widehat{}^2$  & 91.7$\%$ & 0.93 & 0.917 & 0.917 \\
\hline
SAMM & 85.7$\%$ & 0.823& 0.799& 0.81\\
\hline
\end{tabular}
\label{table:statistics}
\end{table}

\section{Results and Discussion}
In this section, first we present the experimental results comparison, then the impact of kernel size on spatio-temporal attention module and the visualization of impact of attention through saliency maps.

\subsection{Experimental Results}
Experiments are conducted by training both the 3D residual network and the proposed 3D residual attention network models. The results are compared in terms of the facial micro-expression recognition accuracy. The performance of the proposed residual attention network is compared with the state-of-the-art hand-crafted methods (HCM) and deep learning methods (DLM) as shown in Table \ref{table:comparison}. The results are computed over three benchmark facial micro-expression datasets, including CASME I, CASME II, and CAS(ME)$^2$ in Table \ref{table:comparison}. Note that the results of competing methods are taken from the corresponding papers. Table \ref{table:comparison} depicts that the proposed MERANet model outperforms the existing hand crafted and deep learning models with a significant margin. As expected, the channel attention and the spatio-temporal attention blocks help the residual network to learn better micro-level spatio-temporal features leading to the enhanced decision making for facial micro-expression recognition. 

The total number of parameters for the original 3D ResNet model is $3,31,67,811$ which, for the proposed 3D MERANet model, is increased to $3,35,47,552$. The proposed model uses $1.72$ GFLOPs whereas original model uses $1.61$ GFLOPs. The number of parameters increased for the proposed model by $3,79,741$, and FLOPs increased by $0.11$G, which is quite small compared to the original 3D ResNet model. So, the overall overhead added by the proposed attention module is negligible in terms of both parameters and computation. Table \ref{table:statistics} represents the results in terms of the Accuracy, Precision, Recall and F1 measures for four benchmark datasets, including $CASME I$, $CASME II$, $CAS(ME)^2$, and $SAMM$. Promising results are obtained using the proposed MERANet model for micro-expression recognition.
The quantitative results provided in Table \ref{table:comparison} and Table \ref{table:statistics} show the efficacy of the proposed attention modules. Table \ref{conf} shows the confusion matrix for the proposed method on the SAMM dataset.

\begin{figure}[!t]
\centering
\includegraphics[width=\columnwidth]{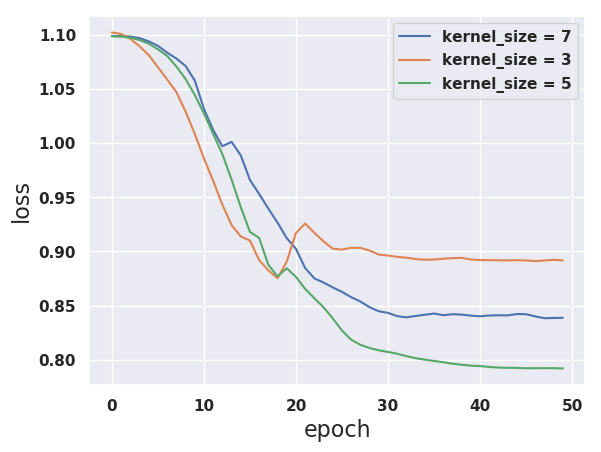}
\caption{The impact of kernel size on the spatio-temporal attention block to prioritize the spatio-temporal information in the intermediate features. The Validation loss with respect to the number of epochs are shown for the kernel sizes of $3$, $5$ and $7$. Medium kernel size is better suited.}
\label{fig:kernelsize}
\end{figure}


\subsection{Impact of Shared Network in Channel Attention}
We now compare the efficacy of shared CNN and shared MLP in channel attention. For this study, we use the $CAS(ME)^{2}$ dataset. In-order to fully capture the interrelationship, the channel descriptor of average pooling $\mathcal{F}_{avg}^{ch}$ and max pooling $\mathcal{F}_{max}^{ch}$ are fed to a shared network. In this subsection, we experimentally verify the proposed two design choices of shared networks, i.e., (a) Channel Attention with shared convolutional neural network (CN-SCNN) and (b) Channel Attention with shared multilayer perceptron (CN-SMLP).
\begin{table}
\centering
\caption{Confusion matrix for the proposed model on SAMM dataset.}
\begin{tabular}{c|c|c|c}
\hline
 & Angry & Happy & Others \\
\hline
Angry & 11 & 0 & 0 \\
\hline
Happy & 0 & 4 & 1 \\
\hline
Other & 1 & 1 & 3 \\
\hline
\end{tabular}
\label{conf}
\end{table}
  
Both the design choices are trained under the same settings. We observe that the residual attention model with shared CNN in the channel attention module work well compared to the model with the shared MLP in the shared network. Shared CNN works on the sliding window concept (which focuses on local regions of interest), whereas shared MLP works on the global feature connections. Due to the aforementioned property, shared CNN goes hand in hand with the attention module by improving the learning capability of the model. The loss curves for both the design choices are plotted in Fig. \ref{fig:validationloss}. From the validation loss curve in Fig. \ref{fig:validationloss}, we observe that the CN-SCNN outperforms the CN-SMLP at every epoch. Model trained with the CN-SCNN module converges faster compared to the model trained with the CN-SMLP. In conclusion, we infer that the shared CNN in the channel attention module is a better design choice compared to the shared MLP.

\subsection{Impact of Kernel Size on Spatio-Temporal Attention}
The performance of spatial-temporal attention block is mainly dependant on the applied convolution operation. In general, convolution layer performance depends on the size of the filter used. The kernel size plays a prominent role in extracting the subtle features. So, we have experimented with different kernel sizes, i.e., $3$, $5$, and $7$ to find the optimal kernel size that boosts the model learning capability. The validation curve is shown in Fig. \ref{fig:kernelsize} having the plots corresponding to each kernel size. It is observed that the smaller kernel size (i.e., small receptive field) doesn't help the model to learn better. In contrast, higher kernel size ($k=7$) increases the computation head, yet it fails to generalize better. From Fig. \ref{fig:kernelsize}, we also infer that a reasonable receptive field is needed for deciding important regions. In the comparison of different convolution kernel sizes, we find the kernel size of $5$ helps the model to learn better micro-expression representation.
\begin{figure}[!t]
\centering
\includegraphics[width=\columnwidth]{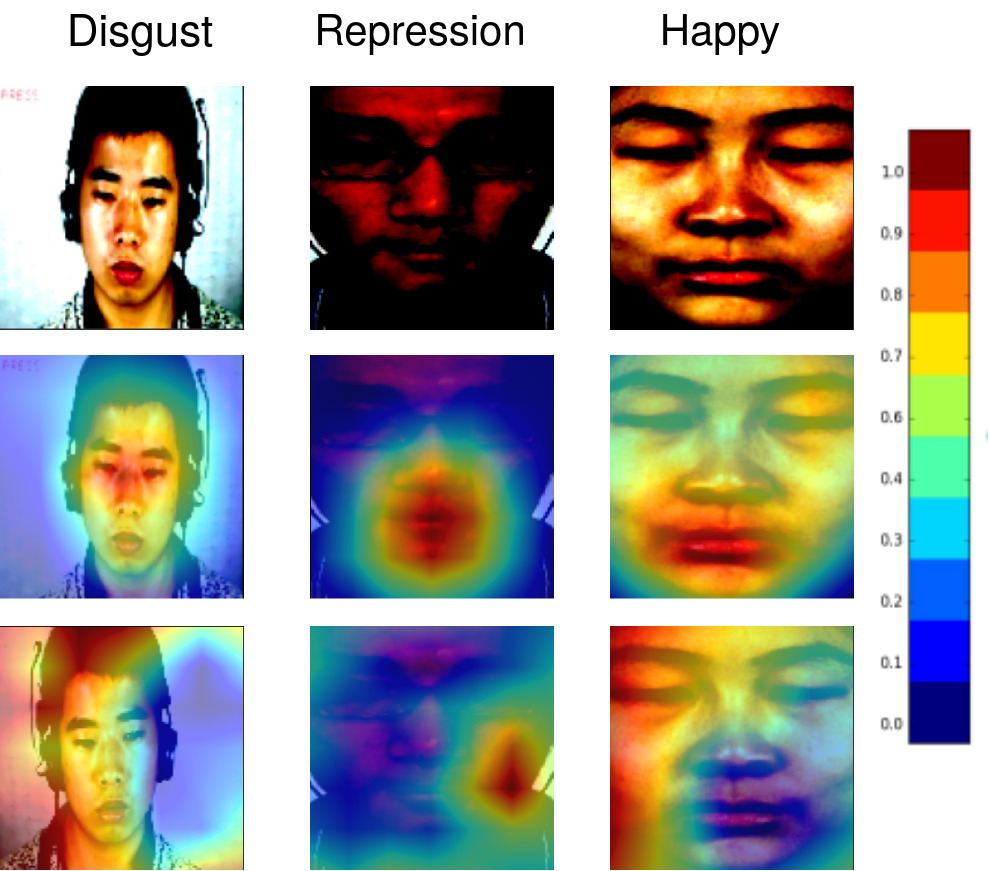}
\caption{Comparison between the proposed MERANet (Attention) with 3D ResNet (No Attention) in terms of the visualizations of the saliency maps provided by the two methods. The $1^{st}$ row depicts the input image, the $2^{nd}$ row contains saliency visualizations from the proposed model and the last row contains the saliency visualizations from the 3D ResNet model. The ground-truth label is shown on the top of each input image.}
\label{fig:saliencymap}
\end{figure}

\subsection{Impact of Attention Through Saliency Map}
In this subsection, we empirically show the effectiveness of our proposed model. For this study, we use all the three datasets.
In order to further understand the behavior of the proposed model, we have used the feature visualization technique proposed in ~\cite{gradcam}. These feature visualizations help us to find which spatial locations trigger the classification outcome, thus providing the insights about the model's understanding of the emotions. The feature visualizations are given in Fig. \ref{fig:saliencymap}. According to \cite{ekman1997face}, an emotion depicting disgust on a subject's face has facial muscle movements centered around the nose of the subject. Similarly, a happy emotion has facial muscle movements centered around the lip region. From Fig. \ref{fig:saliencymap}, we infer that the MERANet focuses on the face regions completely ignoring the background. Furthermore, it gives more weight to facial regions corresponding to an emotion. The saliency maps of 3D ResNet model seem to be very arbitrary by considering the noisy regions in the decision making process. Whereas, the saliency maps for the proposed MERANet model clearly matches the intuition with the general procedure of finding facial expressions and more convincing.

\section{Conclusion}
We have introduced a 3D residual attention based MERANet model for micro-expression recognition. The proposed 3D residual attention module utilizes the benefits of channel attention as well as spatio-temporal attention. These attention blocks help the proposed model to focus over the spatio-temporal features leading to the facial micro-expression recognition. We have conducted the extensive experiments with the proposed model over four benchmark datasets and compared with the state-of-the-art models. The proposed MERANet model outperforms the existing models with a negligible increase in the number of parameters. We also found that the shared CNN is better suitable with channel attention. However, the kernel size of $5$ is better suitable with spatio-temporal attention. The saliency map of the proposed MERANet model is very close to the human perception of the emotions in terms of the facial features contribution towards the identification of the emotions.

{\small
\bibliographystyle{ACM-Reference-Format}
\bibliography{References}


\begin{thebibliography}{44}


\ifx \showCODEN    \undefined \def \showCODEN     #1{\unskip}     \fi
\ifx \showDOI      \undefined \def \showDOI       #1{#1}\fi
\ifx \showISBNx    \undefined \def \showISBNx     #1{\unskip}     \fi
\ifx \showISBNxiii \undefined \def \showISBNxiii  #1{\unskip}     \fi
\ifx \showISSN     \undefined \def \showISSN      #1{\unskip}     \fi
\ifx \showLCCN     \undefined \def \showLCCN      #1{\unskip}     \fi
\ifx \shownote     \undefined \def \shownote      #1{#1}          \fi
\ifx \showarticletitle \undefined \def \showarticletitle #1{#1}   \fi
\ifx \showURL      \undefined \def \showURL       {\relax}        \fi
\providecommand\bibfield[2]{#2}
\providecommand\bibinfo[2]{#2}
\providecommand\natexlab[1]{#1}
\providecommand\showeprint[2][]{arXiv:#2}

\bibitem[\protect\citeauthoryear{Cao, Liu, Yang, Yu, Wang, Wang, Huang, Wang,
  Huang, Xu, et~al\mbox{.}}{Cao et~al\mbox{.}}{2015}]%
        {topdown}
\bibfield{author}{\bibinfo{person}{Chunshui Cao}, \bibinfo{person}{Xianming
  Liu}, \bibinfo{person}{Yi Yang}, \bibinfo{person}{Yinan Yu},
  \bibinfo{person}{Jiang Wang}, \bibinfo{person}{Zilei Wang},
  \bibinfo{person}{Yongzhen Huang}, \bibinfo{person}{Liang Wang},
  \bibinfo{person}{Chang Huang}, \bibinfo{person}{Wei Xu}, {et~al\mbox{.}}}
  \bibinfo{year}{2015}\natexlab{}.
\newblock \showarticletitle{Look and think twice: Capturing top-down visual
  attention with feedback convolutional neural networks}. In
  \bibinfo{booktitle}{\emph{Proceedings of the IEEE international conference on
  computer vision}}. \bibinfo{pages}{2956--2964}.
\newblock


\bibitem[\protect\citeauthoryear{Davison, Lansley, Costen, Tan, and
  Yap}{Davison et~al\mbox{.}}{2018}]%
        {davison2018}
\bibfield{author}{\bibinfo{person}{Adrian~K. Davison}, \bibinfo{person}{Cliff
  Lansley}, \bibinfo{person}{Nicholas Costen}, \bibinfo{person}{Kevin Tan},
  {and} \bibinfo{person}{Moi~Hoon Yap}.} \bibinfo{year}{2018}\natexlab{}.
\newblock \showarticletitle{SAMM: A Spontaneous Micro-Facial Movement Dataset}.
\newblock \bibinfo{journal}{\emph{IEEE Transactions on Affective Computing}}
  \bibinfo{volume}{9} (\bibinfo{year}{2018}), \bibinfo{pages}{116--129}.
\newblock
\urldef\tempurl%
\url{https://doi.org/10.1109/TAFFC.2016.2573832}
\showDOI{\tempurl}


\bibitem[\protect\citeauthoryear{Ekman}{Ekman}{2009}]%
        {ekman}
\bibfield{author}{\bibinfo{person}{Paul Ekman}.}
  \bibinfo{year}{2009}\natexlab{}.
\newblock \showarticletitle{Lie catching and microexpressions}.
\newblock \bibinfo{journal}{\emph{The philosophy of deception}}
  \bibinfo{volume}{1}, \bibinfo{number}{2} (\bibinfo{year}{2009}),
  \bibinfo{pages}{5}.
\newblock


\bibitem[\protect\citeauthoryear{Ekman}{Ekman}{1997}]%
        {ekman1997face}
\bibfield{author}{\bibinfo{person}{Rosenberg Ekman}.}
  \bibinfo{year}{1997}\natexlab{}.
\newblock \bibinfo{booktitle}{\emph{What the face reveals: Basic and applied
  studies of spontaneous expression using the Facial Action Coding System
  (FACS)}}.
\newblock \bibinfo{publisher}{Oxford University Press, USA}.
\newblock


\bibitem[\protect\citeauthoryear{Gan, Liong, Yau, Huang, and Tan}{Gan
  et~al\mbox{.}}{2019}]%
        {liong2018}
\bibfield{author}{\bibinfo{person}{YS Gan}, \bibinfo{person}{Sze-Teng Liong},
  \bibinfo{person}{Wei-Chuen Yau}, \bibinfo{person}{Yen-Chang Huang}, {and}
  \bibinfo{person}{Lit-Ken Tan}.} \bibinfo{year}{2019}\natexlab{}.
\newblock \showarticletitle{Off-apexnet on micro-expression recognition
  system}.
\newblock \bibinfo{journal}{\emph{Signal Processing: Image Communication}}
  \bibinfo{volume}{74} (\bibinfo{year}{2019}), \bibinfo{pages}{129--139}.
\newblock


\bibitem[\protect\citeauthoryear{Glorot and Bengio}{Glorot and Bengio}{2010}]%
        {Xavier}
\bibfield{author}{\bibinfo{person}{Xavier Glorot} {and} \bibinfo{person}{Yoshua
  Bengio}.} \bibinfo{year}{2010}\natexlab{}.
\newblock \showarticletitle{Understanding the difficulty of training deep
  feedforward neural networks}. In \bibinfo{booktitle}{\emph{Proceedings of the
  thirteenth international conference on artificial intelligence and
  statistics}}. \bibinfo{pages}{249--256}.
\newblock


\bibitem[\protect\citeauthoryear{He, Zhang, Ren, and Sun}{He
  et~al\mbox{.}}{2016}]%
        {ResNet}
\bibfield{author}{\bibinfo{person}{Kaiming He}, \bibinfo{person}{Xiangyu
  Zhang}, \bibinfo{person}{Shaoqing Ren}, {and} \bibinfo{person}{Jian Sun}.}
  \bibinfo{year}{2016}\natexlab{}.
\newblock \showarticletitle{Deep residual learning for image recognition}. In
  \bibinfo{booktitle}{\emph{Proceedings of the IEEE conference on computer
  vision and pattern recognition}}. \bibinfo{pages}{770--778}.
\newblock


\bibitem[\protect\citeauthoryear{Hu, Shen, and Sun}{Hu et~al\mbox{.}}{2018}]%
        {squeezeexcite}
\bibfield{author}{\bibinfo{person}{Jie Hu}, \bibinfo{person}{Li Shen}, {and}
  \bibinfo{person}{Gang Sun}.} \bibinfo{year}{2018}\natexlab{}.
\newblock \showarticletitle{Squeeze-and-excitation networks}. In
  \bibinfo{booktitle}{\emph{Proceedings of the IEEE conference on computer
  vision and pattern recognition}}. \bibinfo{pages}{7132--7141}.
\newblock


\bibitem[\protect\citeauthoryear{Huang, Zhao, Hong, Zheng, and
  Pietik{\"a}inen}{Huang et~al\mbox{.}}{2016}]%
        {huang2016spontaneous}
\bibfield{author}{\bibinfo{person}{Xiaohua Huang}, \bibinfo{person}{Guoying
  Zhao}, \bibinfo{person}{Xiaopeng Hong}, \bibinfo{person}{Wenming Zheng},
  {and} \bibinfo{person}{Matti Pietik{\"a}inen}.}
  \bibinfo{year}{2016}\natexlab{}.
\newblock \showarticletitle{Spontaneous facial micro-expression analysis using
  spatiotemporal completed local quantized patterns}.
\newblock \bibinfo{journal}{\emph{Neurocomputing}}  \bibinfo{volume}{175}
  (\bibinfo{year}{2016}), \bibinfo{pages}{564--578}.
\newblock


\bibitem[\protect\citeauthoryear{Itti and Koch}{Itti and Koch}{2001}]%
        {cvatt}
\bibfield{author}{\bibinfo{person}{Laurent Itti} {and}
  \bibinfo{person}{Christof Koch}.} \bibinfo{year}{2001}\natexlab{}.
\newblock \showarticletitle{Computational modelling of visual attention}.
\newblock \bibinfo{journal}{\emph{Nature reviews neuroscience}}
  \bibinfo{volume}{2}, \bibinfo{number}{3} (\bibinfo{year}{2001}),
  \bibinfo{pages}{194--203}.
\newblock


\bibitem[\protect\citeauthoryear{Jelena, Milica, Marina, Angelina, Hasan, and
  Gholamreza}{Jelena et~al\mbox{.}}{[n.d.]}]%
        {grobova2017}
\bibfield{author}{\bibinfo{person}{Grobova Jelena}, \bibinfo{person}{Colovic
  Milica}, \bibinfo{person}{Marjanovic Marina}, \bibinfo{person}{Njegus
  Angelina}, \bibinfo{person}{Demire Hasan}, {and} \bibinfo{person}{Anbarjafari
  Gholamreza}.} \bibinfo{year}{[n.d.]}\natexlab{}.
\newblock \showarticletitle{Automatic Hidden Sadness Detection Using
  Micro-Expressions}. In \bibinfo{booktitle}{\emph{12th IEEE International
  Conference on Automatic Face and Gesture Recognition (FG 2017)}}.
  \bibinfo{publisher}{IEEE}.
\newblock


\bibitem[\protect\citeauthoryear{Khor, See, Phan, and Lin}{Khor
  et~al\mbox{.}}{2018}]%
        {khor2018}
\bibfield{author}{\bibinfo{person}{Huai-Qian Khor}, \bibinfo{person}{John See},
  \bibinfo{person}{Raphael Chung~Wei Phan}, {and} \bibinfo{person}{Weiyao
  Lin}.} \bibinfo{year}{2018}\natexlab{}.
\newblock \showarticletitle{Enriched long-term recurrent convolutional network
  for facial micro-expression recognition}. In \bibinfo{booktitle}{\emph{2018
  13th IEEE International Conference on Automatic Face \& Gesture Recognition
  (FG 2018)}}. IEEE, \bibinfo{pages}{667--674}.
\newblock


\bibitem[\protect\citeauthoryear{Kim, Baddar, Jang, and Ro}{Kim
  et~al\mbox{.}}{2017}]%
        {kim2017}
\bibfield{author}{\bibinfo{person}{Dae~Hoe Kim}, \bibinfo{person}{Wissam~J
  Baddar}, \bibinfo{person}{Jinhyeok Jang}, {and} \bibinfo{person}{Yong~Man
  Ro}.} \bibinfo{year}{2017}\natexlab{}.
\newblock \showarticletitle{Multi-objective based spatio-temporal feature
  representation learning robust to expression intensity variations for facial
  expression recognition}.
\newblock \bibinfo{journal}{\emph{IEEE Transactions on Affective Computing}}
  \bibinfo{volume}{10}, \bibinfo{number}{2} (\bibinfo{year}{2017}),
  \bibinfo{pages}{223--236}.
\newblock


\bibitem[\protect\citeauthoryear{Kim, Baddar, and Ro}{Kim
  et~al\mbox{.}}{2016}]%
        {kim2016micro}
\bibfield{author}{\bibinfo{person}{Dae~Hoe Kim}, \bibinfo{person}{Wissam~J
  Baddar}, {and} \bibinfo{person}{Yong~Man Ro}.}
  \bibinfo{year}{2016}\natexlab{}.
\newblock \showarticletitle{Micro-Expression Recognition with Expression-State
  Constrained Spatio-Temporal Feature Representations}. In
  \bibinfo{booktitle}{\emph{Proceedings of the 2016 ACM on Multimedia
  Conference}}. ACM, \bibinfo{pages}{382--386}.
\newblock


\bibitem[\protect\citeauthoryear{Larochelle and Hinton}{Larochelle and
  Hinton}{2010}]%
        {hinton}
\bibfield{author}{\bibinfo{person}{Hugo Larochelle} {and}
  \bibinfo{person}{Geoffrey~E Hinton}.} \bibinfo{year}{2010}\natexlab{}.
\newblock \showarticletitle{Learning to combine foveal glimpses with a
  third-order Boltzmann machine}. In \bibinfo{booktitle}{\emph{Advances in
  neural information processing systems}}. \bibinfo{pages}{1243--1251}.
\newblock


\bibitem[\protect\citeauthoryear{LeCun, Bengio, and Hinton}{LeCun
  et~al\mbox{.}}{2015}]%
        {lecun2015deep}
\bibfield{author}{\bibinfo{person}{Yann LeCun}, \bibinfo{person}{Yoshua
  Bengio}, {and} \bibinfo{person}{Geoffrey Hinton}.}
  \bibinfo{year}{2015}\natexlab{}.
\newblock \showarticletitle{Deep learning}.
\newblock \bibinfo{journal}{\emph{Nature}} \bibinfo{volume}{521},
  \bibinfo{number}{7553} (\bibinfo{year}{2015}), \bibinfo{pages}{436--444}.
\newblock


\bibitem[\protect\citeauthoryear{Li, Wang, See, and Liu}{Li
  et~al\mbox{.}}{2019}]%
        {li2018flow}
\bibfield{author}{\bibinfo{person}{Jing Li}, \bibinfo{person}{Yandan Wang},
  \bibinfo{person}{John See}, {and} \bibinfo{person}{Wenbin Liu}.}
  \bibinfo{year}{2019}\natexlab{}.
\newblock \showarticletitle{Micro-expression recognition based on 3D flow
  convolutional neural network}.
\newblock \bibinfo{journal}{\emph{Pattern Analysis and Applications}}
  \bibinfo{volume}{22}, \bibinfo{number}{4} (\bibinfo{year}{2019}),
  \bibinfo{pages}{1331--1339}.
\newblock


\bibitem[\protect\citeauthoryear{Li, Yu, Kurihara, and Zhan}{Li
  et~al\mbox{.}}{2018}]%
        {li2018}
\bibfield{author}{\bibinfo{person}{Qiuyu Li}, \bibinfo{person}{Jun Yu},
  \bibinfo{person}{Toru Kurihara}, {and} \bibinfo{person}{Shu Zhan}.}
  \bibinfo{year}{2018}\natexlab{}.
\newblock \showarticletitle{Micro-expression analysis by fusing deep
  convolutional neural network and optical flow}. In
  \bibinfo{booktitle}{\emph{2018 5th International Conference on Control,
  Decision and Information Technologies (CoDIT)}}. IEEE,
  \bibinfo{pages}{265--270}.
\newblock


\bibitem[\protect\citeauthoryear{Liong, Gan, See, Khor, and Huang}{Liong
  et~al\mbox{.}}{2019}]%
        {liong2019}
\bibfield{author}{\bibinfo{person}{Sze-Teng Liong}, \bibinfo{person}{YS Gan},
  \bibinfo{person}{John See}, \bibinfo{person}{Huai-Qian Khor}, {and}
  \bibinfo{person}{Yen-Chang Huang}.} \bibinfo{year}{2019}\natexlab{}.
\newblock \showarticletitle{Shallow triple stream three-dimensional cnn
  (ststnet) for micro-expression recognition}. In
  \bibinfo{booktitle}{\emph{2019 14th IEEE International Conference on
  Automatic Face \& Gesture Recognition (FG 2019)}}. IEEE,
  \bibinfo{pages}{1--5}.
\newblock


\bibitem[\protect\citeauthoryear{Liu, Zhang, Yan, Wang, Zhao, and Fu}{Liu
  et~al\mbox{.}}{2015}]%
        {liu2016main}
\bibfield{author}{\bibinfo{person}{Yong-Jin Liu}, \bibinfo{person}{Jin-Kai
  Zhang}, \bibinfo{person}{Wen-Jing Yan}, \bibinfo{person}{Su-Jing Wang},
  \bibinfo{person}{Guoying Zhao}, {and} \bibinfo{person}{Xiaolan Fu}.}
  \bibinfo{year}{2015}\natexlab{}.
\newblock \showarticletitle{A main directional mean optical flow feature for
  spontaneous micro-expression recognition}.
\newblock \bibinfo{journal}{\emph{IEEE Transactions on Affective Computing}}
  \bibinfo{volume}{7}, \bibinfo{number}{4} (\bibinfo{year}{2015}),
  \bibinfo{pages}{299--310}.
\newblock


\bibitem[\protect\citeauthoryear{Loshchilov and Hutter}{Loshchilov and
  Hutter}{2016}]%
        {Cosine}
\bibfield{author}{\bibinfo{person}{Ilya Loshchilov} {and}
  \bibinfo{person}{Frank Hutter}.} \bibinfo{year}{2016}\natexlab{}.
\newblock \showarticletitle{Sgdr: Stochastic gradient descent with warm
  restarts}.
\newblock \bibinfo{journal}{\emph{arXiv preprint arXiv:1608.03983}}
  (\bibinfo{year}{2016}).
\newblock


\bibitem[\protect\citeauthoryear{Lu, Kpalma, and Ronsin}{Lu
  et~al\mbox{.}}{2018}]%
        {lu2018}
\bibfield{author}{\bibinfo{person}{Hua Lu}, \bibinfo{person}{Kidiyo Kpalma},
  {and} \bibinfo{person}{Joseph Ronsin}.} \bibinfo{year}{2018}\natexlab{}.
\newblock \showarticletitle{Motion descriptors for micro-expression
  recognition}.
\newblock \bibinfo{journal}{\emph{Signal Processing: Image Communication}}
  \bibinfo{volume}{67} (\bibinfo{year}{2018}), \bibinfo{pages}{108--117}.
\newblock


\bibitem[\protect\citeauthoryear{Ma and Yarats}{Ma and Yarats}{2019}]%
        {Uwarmup}
\bibfield{author}{\bibinfo{person}{Jerry Ma} {and} \bibinfo{person}{Denis
  Yarats}.} \bibinfo{year}{2019}\natexlab{}.
\newblock \showarticletitle{On the adequacy of untuned warmup for adaptive
  optimization}.
\newblock \bibinfo{journal}{\emph{arXiv preprint arXiv:1910.04209}}
  (\bibinfo{year}{2019}).
\newblock


\bibitem[\protect\citeauthoryear{Mnih, Heess, Graves, et~al\mbox{.}}{Mnih
  et~al\mbox{.}}{2014}]%
        {humanperception}
\bibfield{author}{\bibinfo{person}{Volodymyr Mnih}, \bibinfo{person}{Nicolas
  Heess}, \bibinfo{person}{Alex Graves}, {et~al\mbox{.}}}
  \bibinfo{year}{2014}\natexlab{}.
\newblock \showarticletitle{Recurrent models of visual attention}. In
  \bibinfo{booktitle}{\emph{Advances in neural information processing
  systems}}. \bibinfo{pages}{2204--2212}.
\newblock


\bibitem[\protect\citeauthoryear{Patel, Hong, and Zhao}{Patel
  et~al\mbox{.}}{2016}]%
        {patel2016selective}
\bibfield{author}{\bibinfo{person}{Devangini Patel}, \bibinfo{person}{Xiaopeng
  Hong}, {and} \bibinfo{person}{Guoying Zhao}.}
  \bibinfo{year}{2016}\natexlab{}.
\newblock \showarticletitle{Selective deep features for micro-expression
  recognition}. In \bibinfo{booktitle}{\emph{2016 23rd international conference
  on pattern recognition (ICPR)}}. IEEE, \bibinfo{pages}{2258--2263}.
\newblock


\bibitem[\protect\citeauthoryear{Peng, Wang, and Chen}{Peng
  et~al\mbox{.}}{2018}]%
        {gesture}
\bibfield{author}{\bibinfo{person}{Min Peng}, \bibinfo{person}{Chongyang Wang},
  {and} \bibinfo{person}{Tong Chen}.} \bibinfo{year}{2018}\natexlab{}.
\newblock \showarticletitle{Attention Based Residual Network for Micro-Gesture
  Recognition}. In \bibinfo{booktitle}{\emph{2018 13th IEEE International
  Conference on Automatic Face \& Gesture Recognition (FG 2018)}}. IEEE,
  \bibinfo{pages}{790--794}.
\newblock


\bibitem[\protect\citeauthoryear{Pfister, Li, Zhao, and
  Pietik{\"a}inen}{Pfister et~al\mbox{.}}{2011}]%
        {pfister2011recognising}
\bibfield{author}{\bibinfo{person}{Tomas Pfister}, \bibinfo{person}{Xiaobai
  Li}, \bibinfo{person}{Guoying Zhao}, {and} \bibinfo{person}{Matti
  Pietik{\"a}inen}.} \bibinfo{year}{2011}\natexlab{}.
\newblock \showarticletitle{Recognising spontaneous facial micro-expressions}.
  In \bibinfo{booktitle}{\emph{2011 international conference on computer
  vision}}. IEEE, \bibinfo{pages}{1449--1456}.
\newblock


\bibitem[\protect\citeauthoryear{Polikovsky, Kameda, and Ohta}{Polikovsky
  et~al\mbox{.}}{2009}]%
        {polikovsky2009facial}
\bibfield{author}{\bibinfo{person}{Senya Polikovsky},
  \bibinfo{person}{Yoshinari Kameda}, {and} \bibinfo{person}{Yuichi Ohta}.}
  \bibinfo{year}{2009}\natexlab{}.
\newblock \showarticletitle{Facial micro-expressions recognition using high
  speed camera and 3D-gradient descriptor}. In \bibinfo{booktitle}{\emph{3rd
  International Conference on Imaging for Crime Detection and Prevention (ICDP
  2009)}}. IET, \bibinfo{pages}{1--6}.
\newblock


\bibitem[\protect\citeauthoryear{Qu, Wang, Yan, and Fu}{Qu
  et~al\mbox{.}}{2016}]%
        {casmesquare}
\bibfield{author}{\bibinfo{person}{Fangbing Qu}, \bibinfo{person}{Su-Jing
  Wang}, \bibinfo{person}{Wen-Jing Yan}, {and} \bibinfo{person}{Xiaolan Fu}.}
  \bibinfo{year}{2016}\natexlab{}.
\newblock \showarticletitle{CAS (ME) 2: A Database of Spontaneous
  Macro-expressions and Micro-expressions}. In
  \bibinfo{booktitle}{\emph{International Conference on Human-Computer
  Interaction}}. Springer, \bibinfo{pages}{48--59}.
\newblock


\bibitem[\protect\citeauthoryear{Reddy, Karri, Dubey, and Mukherjee}{Reddy
  et~al\mbox{.}}{2019}]%
        {reddy2019}
\bibfield{author}{\bibinfo{person}{Sai Prasanna~Teja Reddy},
  \bibinfo{person}{Surya~Teja Karri}, \bibinfo{person}{Shiv~Ram Dubey}, {and}
  \bibinfo{person}{Snehasis Mukherjee}.} \bibinfo{year}{2019}\natexlab{}.
\newblock \showarticletitle{Spontaneous facial micro-expression recognition
  using 3D spatiotemporal convolutional neural networks}. In
  \bibinfo{booktitle}{\emph{2019 International Joint Conference on Neural
  Networks (IJCNN)}}. IEEE, \bibinfo{pages}{1--8}.
\newblock


\bibitem[\protect\citeauthoryear{Selvaraju, Cogswell, Das, Vedantam, Parikh,
  and Batra}{Selvaraju et~al\mbox{.}}{2017}]%
        {gradcam}
\bibfield{author}{\bibinfo{person}{Ramprasaath~R Selvaraju},
  \bibinfo{person}{Michael Cogswell}, \bibinfo{person}{Abhishek Das},
  \bibinfo{person}{Ramakrishna Vedantam}, \bibinfo{person}{Devi Parikh}, {and}
  \bibinfo{person}{Dhruv Batra}.} \bibinfo{year}{2017}\natexlab{}.
\newblock \showarticletitle{Grad-cam: Visual explanations from deep networks
  via gradient-based localization}. In \bibinfo{booktitle}{\emph{Proceedings of
  the IEEE international conference on computer vision}}.
  \bibinfo{pages}{618--626}.
\newblock


\bibitem[\protect\citeauthoryear{Shreve, Godavarthy, Goldgof, and
  Sarkar}{Shreve et~al\mbox{.}}{2011}]%
        {shreve2011macro}
\bibfield{author}{\bibinfo{person}{Matthew Shreve}, \bibinfo{person}{Sridhar
  Godavarthy}, \bibinfo{person}{Dmitry Goldgof}, {and} \bibinfo{person}{Sudeep
  Sarkar}.} \bibinfo{year}{2011}\natexlab{}.
\newblock \showarticletitle{Macro-and micro-expression spotting in long videos
  using spatio-temporal strain}. In \bibinfo{booktitle}{\emph{IEEE
  International Conference on Face and Gesture (FG 2011)}}. IEEE,
  \bibinfo{pages}{51--56}.
\newblock


\bibitem[\protect\citeauthoryear{S{\o}nderby, S{\o}nderby, Maal{\o}e, and
  Winther}{S{\o}nderby et~al\mbox{.}}{2015}]%
        {Transformer}
\bibfield{author}{\bibinfo{person}{S{\o}ren~Kaae S{\o}nderby},
  \bibinfo{person}{Casper~Kaae S{\o}nderby}, \bibinfo{person}{Lars Maal{\o}e},
  {and} \bibinfo{person}{Ole Winther}.} \bibinfo{year}{2015}\natexlab{}.
\newblock \showarticletitle{Recurrent spatial transformer networks}.
\newblock \bibinfo{journal}{\emph{arXiv preprint arXiv:1509.05329}}
  (\bibinfo{year}{2015}).
\newblock


\bibitem[\protect\citeauthoryear{Takalkar and Xu}{Takalkar and Xu}{2017}]%
        {takalkar2017}
\bibfield{author}{\bibinfo{person}{Madhumita~A Takalkar} {and}
  \bibinfo{person}{Min Xu}.} \bibinfo{year}{2017}\natexlab{}.
\newblock \showarticletitle{Image based facial micro-expression recognition
  using deep learning on small datasets}. In \bibinfo{booktitle}{\emph{2017
  international conference on digital image computing: techniques and
  applications (DICTA)}}. IEEE, \bibinfo{pages}{1--7}.
\newblock


\bibitem[\protect\citeauthoryear{Verma, Vipparthi, Singh, and Murala}{Verma
  et~al\mbox{.}}{2019}]%
        {Verma_2020}
\bibfield{author}{\bibinfo{person}{Monu Verma}, \bibinfo{person}{Santosh~Kumar
  Vipparthi}, \bibinfo{person}{Girdhari Singh}, {and}
  \bibinfo{person}{Subrahmanyam Murala}.} \bibinfo{year}{2019}\natexlab{}.
\newblock \showarticletitle{LEARNet: Dynamic imaging network for micro
  expression recognition}.
\newblock \bibinfo{journal}{\emph{IEEE Transactions on Image Processing}}
  \bibinfo{volume}{29} (\bibinfo{year}{2019}), \bibinfo{pages}{1618--1627}.
\newblock


\bibitem[\protect\citeauthoryear{Wang, Peng, Bi, and Chen}{Wang
  et~al\mbox{.}}{2020}]%
        {wang2019}
\bibfield{author}{\bibinfo{person}{Chongyang Wang}, \bibinfo{person}{Min Peng},
  \bibinfo{person}{Tao Bi}, {and} \bibinfo{person}{Tong Chen}.}
  \bibinfo{year}{2020}\natexlab{}.
\newblock \showarticletitle{Micro-attention for micro-expression recognition}.
\newblock \bibinfo{journal}{\emph{Neurocomputing}}  \bibinfo{volume}{410}
  (\bibinfo{year}{2020}), \bibinfo{pages}{354--362}.
\newblock


\bibitem[\protect\citeauthoryear{Wang, Jiang, Qian, Yang, Li, Zhang, Wang, and
  Tang}{Wang et~al\mbox{.}}{2017}]%
        {RAN}
\bibfield{author}{\bibinfo{person}{Fei Wang}, \bibinfo{person}{Mengqing Jiang},
  \bibinfo{person}{Chen Qian}, \bibinfo{person}{Shuo Yang},
  \bibinfo{person}{Cheng Li}, \bibinfo{person}{Honggang Zhang},
  \bibinfo{person}{Xiaogang Wang}, {and} \bibinfo{person}{Xiaoou Tang}.}
  \bibinfo{year}{2017}\natexlab{}.
\newblock \showarticletitle{Residual attention network for image
  classification}. In \bibinfo{booktitle}{\emph{Proceedings of the IEEE
  conference on computer vision and pattern recognition}}.
  \bibinfo{pages}{3156--3164}.
\newblock


\bibitem[\protect\citeauthoryear{Woo, Park, Lee, and So~Kweon}{Woo
  et~al\mbox{.}}{2018}]%
        {CBAM}
\bibfield{author}{\bibinfo{person}{Sanghyun Woo}, \bibinfo{person}{Jongchan
  Park}, \bibinfo{person}{Joon-Young Lee}, {and} \bibinfo{person}{In
  So~Kweon}.} \bibinfo{year}{2018}\natexlab{}.
\newblock \showarticletitle{Cbam: Convolutional block attention module}. In
  \bibinfo{booktitle}{\emph{Proceedings of the European conference on computer
  vision (ECCV)}}. \bibinfo{pages}{3--19}.
\newblock


\bibitem[\protect\citeauthoryear{Xie, Girshick, Doll{\'a}r, Tu, and He}{Xie
  et~al\mbox{.}}{2017}]%
        {ResNeXt}
\bibfield{author}{\bibinfo{person}{Saining Xie}, \bibinfo{person}{Ross
  Girshick}, \bibinfo{person}{Piotr Doll{\'a}r}, \bibinfo{person}{Zhuowen Tu},
  {and} \bibinfo{person}{Kaiming He}.} \bibinfo{year}{2017}\natexlab{}.
\newblock \showarticletitle{Aggregated residual transformations for deep neural
  networks}. In \bibinfo{booktitle}{\emph{Proceedings of the IEEE conference on
  computer vision and pattern recognition}}. \bibinfo{pages}{1492--1500}.
\newblock


\bibitem[\protect\citeauthoryear{Yan, Li, Wang, Zhao, Liu, Chen, and Fu}{Yan
  et~al\mbox{.}}{2014}]%
        {Yan2014CASMEIA}
\bibfield{author}{\bibinfo{person}{Wen-Jing Yan}, \bibinfo{person}{Xiaobai Li},
  \bibinfo{person}{Su-Jing Wang}, \bibinfo{person}{Guoying Zhao},
  \bibinfo{person}{Yong-Jin Liu}, \bibinfo{person}{Yu-Hsin Chen}, {and}
  \bibinfo{person}{Xiaolan Fu}.} \bibinfo{year}{2014}\natexlab{}.
\newblock \showarticletitle{CASME II: An improved spontaneous micro-expression
  database and the baseline evaluation}.
\newblock \bibinfo{journal}{\emph{PloS one}} \bibinfo{volume}{9},
  \bibinfo{number}{1} (\bibinfo{year}{2014}), \bibinfo{pages}{e86041}.
\newblock


\bibitem[\protect\citeauthoryear{Yan, Wu, Liu, Wang, and Fu}{Yan
  et~al\mbox{.}}{2013}]%
        {casme}
\bibfield{author}{\bibinfo{person}{Wen-Jing Yan}, \bibinfo{person}{Qi Wu},
  \bibinfo{person}{Yong-Jin Liu}, \bibinfo{person}{Su-Jing Wang}, {and}
  \bibinfo{person}{Xiaolan Fu}.} \bibinfo{year}{2013}\natexlab{}.
\newblock \showarticletitle{CASME database: a dataset of spontaneous
  micro-expressions collected from neutralized faces}. In
  \bibinfo{booktitle}{\emph{2013 10th IEEE international conference and
  workshops on automatic face and gesture recognition (FG)}}. IEEE,
  \bibinfo{pages}{1--7}.
\newblock


\bibitem[\protect\citeauthoryear{Zagoruyko and Komodakis}{Zagoruyko and
  Komodakis}{2016a}]%
        {payattention}
\bibfield{author}{\bibinfo{person}{Sergey Zagoruyko} {and}
  \bibinfo{person}{Nikos Komodakis}.} \bibinfo{year}{2016}\natexlab{a}.
\newblock \showarticletitle{Paying more attention to attention: Improving the
  performance of convolutional neural networks via attention transfer}.
\newblock \bibinfo{journal}{\emph{arXiv preprint arXiv:1612.03928}}
  (\bibinfo{year}{2016}).
\newblock


\bibitem[\protect\citeauthoryear{Zagoruyko and Komodakis}{Zagoruyko and
  Komodakis}{2016b}]%
        {wideresnet}
\bibfield{author}{\bibinfo{person}{Sergey Zagoruyko} {and}
  \bibinfo{person}{Nikos Komodakis}.} \bibinfo{year}{2016}\natexlab{b}.
\newblock \showarticletitle{Wide residual networks}.
\newblock \bibinfo{journal}{\emph{arXiv preprint arXiv:1605.07146}}
  (\bibinfo{year}{2016}).
\newblock


\bibitem[\protect\citeauthoryear{Zhao and Pietikainen}{Zhao and
  Pietikainen}{2007}]%
        {zhao2007dynamic}
\bibfield{author}{\bibinfo{person}{Guoying Zhao} {and} \bibinfo{person}{Matti
  Pietikainen}.} \bibinfo{year}{2007}\natexlab{}.
\newblock \showarticletitle{Dynamic texture recognition using local binary
  patterns with an application to facial expressions}.
\newblock \bibinfo{journal}{\emph{IEEE transactions on pattern analysis and
  machine intelligence}} \bibinfo{volume}{29}, \bibinfo{number}{6}
  (\bibinfo{year}{2007}), \bibinfo{pages}{915--928}.
\newblock


\end{thebibliography}
}

\end{document}